%% file: main.tex
\documentclass[conference]{IEEEtran}
\usepackage{times}
\usepackage[utf8]{inputenc}
\IEEEoverridecommandlockouts
\usepackage[ruled, linesnumbered]{algorithm2e}
\usepackage{color}
\usepackage{fancyhdr}
\usepackage{subcaption}
\usepackage{xcolor}
\usepackage{graphicx}
\usepackage{amsmath,amssymb,amsfonts}
\usepackage{draftwatermark}
\usepackage{amsthm}
\usepackage{comment}
\usepackage{algorithmic}
\usepackage{textcomp}
\usepackage{tabularx}
\usepackage{cancel}
\usepackage[normalem]{ulem}
\usepackage{url}
\usepackage[
maxcitenames=1
]{biblatex}
\def\BibTeX{{\rm B\kern-.05em{\sc i\kern-.025em b}\kern-.08em
    T\kern-.1667em\lower.7ex\hbox{E}\kern-.125emX}}

\author{
    \IEEEauthorblockN{Sangsu Lee\IEEEauthorrefmark{2}, Xi Zheng\IEEEauthorrefmark{3}, Jie Hua\IEEEauthorrefmark{2}, Haris Vikalo\IEEEauthorrefmark{2}, Christine Julien\IEEEauthorrefmark{2}}
    \IEEEauthorblockA{\IEEEauthorrefmark{2}Department of Electrical and Computer Engineering, University of Texas at Austin
    \\\{sethlee, mich94hj, c.julien\}@utexas.edu, hvikalo@ece.utexas.edu}
    \IEEEauthorblockA{\IEEEauthorrefmark{3}Department
of Computing, Macquarie University, james.zheng@mq.edu.au}
}

\title{Opportunistic Federated Learning:\\An Exploration of 
Egocentric Collaboration for Pervasive Computing Applications}
\newcommand{\data}{\mathcal{D}}
\newcommand{\device}{\mathcal{C}}
\newcommand{\lab}{\mathcal{L}}
\newcommand{\goal}{\mathcal{G}}

\usepackage{draftcopy}		
\usepackage{ifthen}

\newboolean{authnotes}
 \setboolean{authnotes}{false}
\ifthenelse{\boolean{authnotes}}
{
\pagestyle{plain}
\newcommand{\cj}[1]{\footnote{\color{magenta}{\bf Christine: #1}}}

\newcommand{\jh}[1]{\footnote{\color{cyan}{\bf Jie: #1}}}
\newcommand{\jz}[1]{\footnote{\color{red}{\bf James: #1}}}

\newcommand{\tocheckJZ}[1]{{\color{red} #1}}
\newcommand{\tbc}[1]{{\color{blue} #1}} 

\usepackage{draftwatermark}
\SetWatermarkLightness{0.95}
\SetWatermarkScale{1}
}
{
\newcommand{\cj}[1]{}
\newcommand{\jh}[1]{}
\newcommand{\jz}[1]{}

\newcommand{\tbc}[1]{#1}

\SetWatermarkText{}
}

\begin{document}
\maketitle

\begin{abstract}
Pervasive computing applications commonly involve user's personal smartphones collecting data to influence application behavior. Applications are often backed by models that learn from the user's experiences to provide personalized and responsive behavior. While models are often pre-trained on massive datasets, {\em federated learning} has gained attention for its ability to train globally shared models on users' private data without requiring the users to share their data directly. However, federated learning requires devices to collaborate via a central server, under the assumption that all users desire to learn the same model. We define a new approach, {\em opportunistic federated learning}, in which individual devices belonging to different users seek to learn robust models that are {\em personalized} to their user's own experiences. However, instead of learning in isolation, these models opportunistically incorporate the learned experiences of other devices they encounter opportunistically. In this paper, we explore the feasibility and limits of such an approach, culminating in a framework that supports encounter-based pairwise collaborative learning. The use of our opportunistic encounter-based learning amplifies the performance of personalized learning while resisting overfitting to encountered data.
\end{abstract}


\section{Introduction}
\input{introduction}

\section{Motivation and Related Work}\label{sec:related}
\input{relatedwork}

\section{Opportunistic Federated Learning}\label{sec:overview}
\input{overview}


\section{Evaluation}\label{sec:evaluation}
\input{empiricalSupport}

\section{Conclusions and Future Work}\label{sec:futurework}
\input{conclusion}
\input{futureworks}
\printbibliography

\end{document}

%% file: introduction.tex
Smartphones, wearable devices, and other devices that fill pervasive computing environments are imbued with increasingly complex sensing, computation, and communication. However, applications still primarily rely on centrally located servers  to support building and executing predictive models for real-time interactions. In this paper, we define {\em opportunistic federated learning}, which explores the potential for device-to-device collaboration to build expressive, accurate, and personalized models for use in pervasive computing applications.

{\bf The Setting.} We consider pervasive computing applications that rely on models for classification, recommendation, or prediction. Examples include predicting the next word a smartphone user might type, classifying objects in a captured image, or predicting whether a captured photo is likely to be shared on social media. The training data for these models is {\em crowdsourced} -- it is generated by a distributed set of independent devices. The derived models are potentially {\em personal}, both with respect to the fact that their outputs may be tailored to an individual and to the fact that the training data may be privileged, private, or proprietary. The goal is not for devices to converge to a single global model but rather for them to selectively collaborate to construct personalized goal models. The devices comprising pervasive computing environments are generally commodity smartphones, with on-board computing, storage, and wireless communication (e.g., WiFi and Bluetooth). They are resource constrained in energy, computation, and communication, yet they are capable of communicating directly with one another through opportunistic encounters.

{\bf Contributions.} We introduce {\em opportunistic federated learning} and a novel approach to model sharing that we term {\em opportunistic momentum}. Devices belonging to individuals are bootstrapped with an initial model, which they {\em personalize} based on their experiences, as represented by the data collected by the device. When a device (the {\em learner}) encounters another device opportunistically (the {\em neighbor}), it uses a summary of the neighbor's available data to determine whether asking for learning support is (1)~beneficial, based on the similarity (or dissimilarity) in their training data and learning goals and (2)~feasible, based on the expected duration of the potentially fleeting encounter. The latter is predicted based on real-time information about mobility patterns and other context. If model collaboration is likely beneficial and feasible, the devices opportunistically exchange model gradients to generate a new local model for the learner. We present our approach as an egocentric one; of course it is possible that both devices in a pair can benefit from the assistance of the other, and an encounter may be used to support both participants as learners, depending on communication and computation constraints. Concretely, the paper's novel contributions are:
\begin{itemize}
    \item We introduce {\em opportunistic federated learning} as an architectural pattern for learning from encounters and {\em opportunistic momentum} as an algorithmic tool for incorporating experiences of others.
    \item We examine the feasibility of opportunistic federated learning with respect to realistic differences in data distributions in pervasive computing networks.
    \item We examine the practicality of opportunistic federated learning with respect to the duration of encounters in pervasive computing applications.
\end{itemize}
In the long term, opportunistic federated learning will be one piece of a larger ecosystem in which cloud, edge, and opportunistic interactions are used in concert based on the instantaneous network conditions and application requirements.

%% file: relatedwork.tex
{\bf Motivating Applications.}
Pervasive computing is teeming with applications that benefit from machine learning but for which  training data is inherently private. These applications are often best served by {\em personalized} models. We focus on problems where the training task is {\em self-labeling}, including applications like keyboards that predict the next emoji the user will select based on the text they have typed~\cite{ramaswamy2019federated}, activity recognition on smartphones~\cite{rokni2018personalized}, or predictions of the popularity of content in social networks~\cite{zhang2019image}. Self-labeling data makes it possible for devices to generate training data on-the-fly as part of a user's normal interaction with their device.

Different devices may have different experiences and generate widely varying  data. We capture this diversity as a skew in the devices' {\em data label distributions}, i.e., the fraction of each label that a device ``sees''~\cite{li2020lotteryfl, verma2019approaches}. A car driven primarily on local roads may have very few images of semi-trucks in its data set, while a long-haul truck may have few samples of bicycles or children playing. An pedestrian application that recognizes landmarks might collect images of trees, mailboxes, and stop signs in the suburbs; the same application in a city center might see traffic lights, large buildings, and road signs. In emoji prediction, the emojis used by a teenager are likely to be very different from those used by a middle-aged adult.

Different devices may also have different {\em goal distribution}, i.e., the subset of labels the device wants to be able to classify correctly. While some goal distributions may be the entire set of labels, many goal distributions will be a subset of the label space. An object recognition system for vehicles may need to learn the entire label set for safety reasons. A landmark recognition system's goal distribution may be identical to its data distribution. And a user's emoji goal distribution may include any emoji used by others of a similar demographic. Mobile devices are capable of on-device training but they do not want to share raw data. However, they may learn from neighboring devices that they encounter opportunistically. The characteristics that underlie our target applications are:
\begin{itemize}
    \item {\em data distribution diversity}: the data one device encounters often differs from the data other devices encounter
    \item {\em goal set diversity}: two devices' goal distributions may be very different and may differ from the data distributions
    \item {\em encounter benefit}: devices benefit from opportunistic collaboration, but the benefits depend on overlaps between the devices' goal and data label distributions
    \item {\em data privacy}: devices are not willing to share their raw data with other devices they encounter opportunistically
\end{itemize}

{\bf Background and Related Work.} As the capabilities of devices and the desire for data privacy have increased, federated and decentralized learning have emerged. We take for granted that deep learning has already moved into the pervasive computing world and focus here on efforts that go beyond applying inference to also enable some form of learning within the pervasive computing devices.

In federated learning, devices collaborate to construct a global model in a way that enables each individual device to maintain the privacy of its own data~\cite{konevcny2016federated, mcmahan2017communication}. Classically, a central coordinator orchestrates the process by delivering a model to each remote device, collecting and aggregating devices' contributions to training that model, then generating and distributing an updated model to continue the process. There are many applications of federated learning in pervasive computing. Wake word detection, also known as keyword spotting in smart home voice assistants, can use federated learning, protecting the potentially private audio data collected at users' end devices~\cite{leroy2019federated}. One of the most classic federated learning applications is next word prediction on a mobile device keyboard~\cite{hard2018federated}. Still other applications have explored on-device image classification and image processing~\cite{xiong2019antnets}. 

Recent work has also explored distributing the federated learning task across a hierarchical edge network~\cite{2020arXiv200709511H} or selecting a coordinator from among a set of fog nodes~\cite{zhang20:achieving}. These approaches are driven by a single coordinator and aim to learn a single global model. This differs from our goal, in which the effort is  distributed and opportunistic, with each device operating egocentrically to improve its own model.

Related efforts in decentralized and personalized learning remove the coordinator. Some approaches frame the goal as a distributed consensus problem and rely solely on peer-to-peer interactions to disseminate model updates~\cite{colin2016gossip, iutzeler2013asynchronous}.  
Others personalize federated learning on mobile devices~\cite{jiang2019improving}, even when users are expected to have diverse learning goals~\cite{wang2019federated}. These efforts personalize local models that optimize a client's model against its own dataset, while still contributing to the training of a shared global model. Others have used collaboration among devices to improve local learning~\cite{vanhaesebrouck2017decentralized}; these approaches place significant constraints on collaboration or make strong assumptions about predictable contacts.

In simultaneously performing federated learning for a global goal and personalization for a local one, existing work has each client solve an optimization problem over its local data using a hyper-parameter that specifies the trade-off between the accuracy of the global and local models~\cite{deng2020adaptive}. Alternatively, meta-learning can be used to adapt a global model to a local dataset~\cite{fallah2020personalized}. These efforts are still based on a traditional federated learning backbone with a central server. Finally, recent work clusters clients with similar data distributions~\cite{mansour2020three} or similar local updates to the global model~\cite{briggs2020federated} and trains a group model. Though these approaches provide some improvement with respect to both the global and local models~\cite{ghosh2020efficient}, they are less applicable to pervasive computing, where clients are moving and their communications is opportunistic.

Even in federated learning, sharing model parameters or gradients potentially reveals something about private data, and  efforts exist to attempt to reverse engineer these abstractions and recover sensitive information~\cite{zhu2019deep}. In practice, however, these techniques are limited, and the working consensus is that federated learning provides increased privacy relative to centralizing raw data~\cite{huang2020instahide}. In addition, privacy is not the only reason to employ federated learning;  sharing  model gradients can incur reduced communication costs relative to directly sharing the raw data~\cite{konevcny2016federated}.

%% file: overview.tex
We focus on applications that rely on deep learning models for prediction or classification tasks in which each device personalizes the model for its own use~\cite{dinh2020personalized, jiang2019improving, wang2019federated}. In contrast to prior work, we examine the benefits of incorporating opportunistic collaboration. Similarly to federated learning~\cite{mcmahan2017communication}, collaborating peers desire to protect their raw data and instead share only snapshots of learned models.

\begin{figure}[t]
\centering
\includegraphics[width=.85\linewidth]{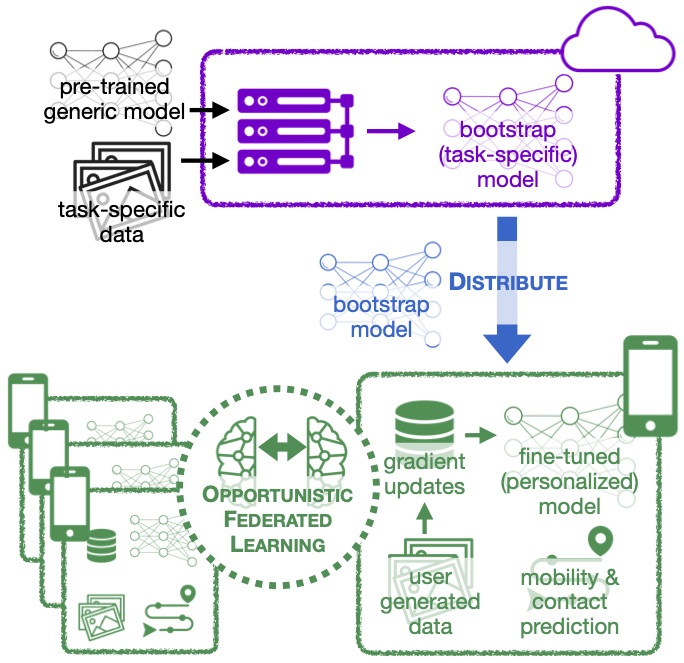}
\caption{\small System Architecture. The process optionally starts with a pre-trained model, which is specialized using task-specific data into the {\em bootstrap model} that is distributed to participating devices. Devices continuously fine-tunes a personalized model using locally collected data. Each device also uses on-device mobility and contact prediction to help decide when to engage in {\em opportunistic federated learning} exchanges with encountered devices.} 
\label{fig:arch}
\end{figure}

Fig.~\ref{fig:arch} shows an overview of our framework. We start with a pre-trained generic model that we tailor using task-specific data to create a {\em bootstrap model} that is distributed to participating devices.  
Devices are controlled and carried by individual users, and they generate or collect (labeled) data. This local data is used to continuously fine-tune (i.e., personalize) the local model. When a device (the ``learner'') encounters another device (``the neighbor''), it may request the neighbor to perform a round of training on the learner's model using the neighbor's data. The learner sends its model parameters to the neighbor; the neighbor trains the model with its local data and returns the gradients, which the learner incorporates into its personal model. 
Though we assume that neighboring devices will participate, there is plenty of work on incentivizing collaboration in opportunistic environments~\cite{kang2019incentive,khan2019federated}, which we can adopt in our scenarios in future work.

We describe the procedure for deciding whether to initiate this process in more detail below; as a preview, it is based on two inputs: (1)~a comparison of the neighbor's data label distribution with the learner's goal distribution and (2)~a prediction of the expected duration of the encounter. For the second, we rely on a long history of prior work in mobility and contact prediction~\cite{zhou17:predicting, chen2016contact}.

\subsection{Data Label Distributions and Goal Distributions}
We assume a set of $N$ devices $\boldsymbol{C} = \{\device_1, \ldots, \device_i, \ldots, \device_N\}$, each with its own local data set, $\data_i$. In our example applications, these data sets are generated when the device takes images of its surroundings, shares images to social media, or sends text messages with emojis. Each sample is associated with a label (e.g., a photo may be labeled with a landmark or object within it; another photo may be labeled with whether it was shared on social media; a sequence of words may be labeled with the emoji that follows them). We capture device $\device_i$'s {\em data label distribution} ($\lab_i$) as the relative frequency of each label within the local data set. The goal of opportunistic federated learning is to learn a local model for some task, i.e., to learn a model that can correctly label a novel input. Each device has a {\em goal distribution} ($\goal_i$) of labels that it desires for its model to be successful at classifying. It is common that a given device's goal and data label distribution differ, and any two given devices may have different goal distributions.

\subsection{Opportunistic Federated Learning}
Each device $\device_i$ constructs its own local model, $w_i$. Because the local model changes over time based on the device's local training and encounters, we indicate a local clock $t_i$ associated with $\device_i$'s model ($w^{t_i}_i$); we increment $t_i$ every time $\device_i$'s local model is updated. $\device_i$'s bootstrap model is $w^{0}_i$. Opportunistic federated learning updates the local model based on a combination of the device's local data and gradients obtained from encounters in order to optimize the local model for the goal distribution $\goal_i$. Formally:
\begin{equation}
    \min_{\boldsymbol{E_i}}\bigg\{\sum_{t_i=0}^{|\boldsymbol{E_i}|}\ell(w^{t_i}_i;\data_{\goal_i})\bigg\},
\end{equation}

where $\boldsymbol{E_i}$ refers to the set of encounters that $\device_i$ has, and $\data_{\goal_i}$ is a hypothetical data set with a data label distribution that satisfies the goal distribution $\goal_i$. Intuitively, we strive to minimize the loss of a model that is learned from incorporating training across the encounters in $\boldsymbol{E_i}$. 

The workflow is shown in Algorithm~\ref{alg:req}. All devices participate in  continuous neighbor discovery~\cite{julien17:blend, kindt2019optimal}, through which they  opportunistically discover nearby devices. The framework relies on three things: a similarity metric, a contact duration prediction, and a mapping of labels to learned gradients. 

The algorithm first computes the similarity between the learner's goal distribution and the neighbor's data label distribution to determine whether the encounter will provide a useful learning opportunity. While there are many sophisticated metrics for similarity, our distributions are not wildly diverse, so we opt for a relatively simple metric. Specifically:
\begin{equation}
    \it{sim}(\mathcal{P}_1, \mathcal{P}_2) = \sum_{l\in L} \min ( \mathcal{P}_1(l), \mathcal{P}_2(l) ).
    \label{eq:sim}
\end{equation}

Intuitively, if a label appears in both distributions, the sum includes the minimum frequency of that label in the two distributions. If two distributions have no labels in common, the similarity score will be 0; if the two distributions are identical, the similarity score will be 1. 

\begin{algorithm}[t]
\caption{Opportunistic Federated Learning}
\label{alg:req}
\DontPrintSemicolon
{\small
\SetKwProg{onDiscover}{Function {\sc onDiscover:}}{}{end}

$\it{sim}(\mathcal{P}_1,\mathcal{P}_2)$: similarity metric for label distributions\\
$\it{contact}(t_i, \device_i, \device_j)$: predicted duration of encounter with device $j$ at time $t_i$\\
$\Gamma[\lab]$: maps a subset of labels to the most recent encounter gradient trained on those labels\\
\onDiscover{$\lab_j$}{
    \If{$\it{sim}(\goal_i, \lab_j) > \tau$}{
        $w^{\prime} = w_i^{t_i}$\\
        \For{$r\gets0$ \KwTo $\rho$}{
            request computation of $\nabla\ell(w^{\prime}; \data_j)$\\
            $\Gamma[\lab_j] \gets\nabla\ell(w^{\prime}; \data_j)$\\
            $w^{\prime}\gets$ \sc{aggregateGradients}\\
        }
        $w_i^{t+1} \gets w^{\prime}$
    }
}
}
\end{algorithm}

Algorithm~\ref{alg:req} also relies on a predicted contact duration between two devices $\mathcal{C}_i$ and $\mathcal{C}_j$ at time $t_i$. This is provided by the underlying system, based on a system-level algorithm on each device that uses the device's context information to determine a likely length of contact. The specific implementation is outside the scope of this paper; we assume an off-the-shelf method to estimate  contact duration~\cite{zhou17:predicting, chen2016contact}.

Thirdly, Algorithm~\ref{alg:req} uses a data structure, $\Gamma$, that maps subsets of the label space (the keys) to gradients learned from encounters (the values). This structure is initially empty, but as the learner encounters and interacts with neighbors, it fills up with mappings from each label subset to the most recent gradients learned on that label subset. For instance, if the neighbor's data label distribution contains labels ${A, B, C}$, the learner will map the subset ${A, B, C}$ to the final gradient returned from the learner in Algorithm~\ref{alg:req}. 

When a device discovers a neighbor, the devices immediately exchange data label distributions. Each device independently executes the {\sc onDiscover} function that starts on line 4 of Algorithm~\ref{alg:req}. The local device (the learner) compares the neighbor's data label distribution to its own goal distribution using Equation~\ref{eq:sim}. If the similarity is greater than a threshold ($\tau$), the learner decides to engage with the neighbor to perform a session of remote training. This session comprises a customizable $\rho$ number of rounds; $\rho$ may be dictated by the task and underlying model, by the expected duration of the contact, or by some combination. If the pair of devices needs to split the duration of the encounter to perform the exchange in both directions, $\rho$ might also be limited and negotiated.

Each round within a session (lines 8-10 in Algorithm~\ref{alg:req}) has two steps. First, the learner ($i$) requests remote learning from the neighbor ($j$). The learner sends the neighbor a copy of its current model by sending a summary of the model parameters. The neighbor loads the model and uses its own local data to compute the gradient $\nabla\ell(w^{\prime}; \data_j)$, which it returns to the learner. The learner stores the returned gradient in the $\Gamma$ data structure, mapped to the label set $\mathcal{L}_j$. If a mapping to $\mathcal{L}_j$ already exists, it is replaced. The learner applies one of a suite of gradient aggregation algorithms (described below), including a round of training on its own local data. These actions update the local model, which is used in any remaining rounds (i.e., the updated model is sent to the neighbor to repeat steps 8-10 in Algorithm~\ref{alg:req}). When the algorithm has completed $\rho$ rounds, the learner's local model is updated, and the device is ready for the next encounter.

\subsection{Aggregating Encounters}

We implement two general options for filling in the {\sc aggregateGradients} function in line 10 of Algorithm~\ref{alg:req} to update the local model. We refer to the first of these as {\em greedy aggregation}. Simply put, greedy aggregation directly averages in the gradients learned by the neighbor after incorporating one round of local learning. Formally, the update to the model $w^{\prime}$ in line 10 of Algorithm~\ref{alg:req} is computed as:
\begin{equation}
    \frac{\omega_{(\lab_i, \goal_i)}(\nabla\ell(w^{\prime}; \data_i)) + \omega_{(\lab_j, \goal_i)}(\nabla\ell(w^{\prime}; \data_j))}{\omega_{(\lab_i, \goal_i)}+ \omega_{(\lab_j, \goal_i)}},
    \label{eq:greedy}
\end{equation}

The two gradients (the local one and the one from the neighbor) are both weighted with respect to the similarity of the corresponding data label distribution with the goal distribution. Building on existing work that similarly uses weights to address unbalanced data~\cite{deng2020adaptive}, the weights are: 
\begin{equation}
    \omega_{(\lab,\goal)} = \exp(-\lambda\times(1-\it{sim}(\goal, \lab))),
    \label{eq:weights}
\end{equation}
where $\lambda$ reflects how a model is prone to overfitting to a dataset that is small relative to the total number of labels. It can also be interpreted as a model's preference for a highly balanced dataset. For our experiments, we obtained $\lambda$ empirically 
by running a series of personalization rounds using subsets of the training data used for the bootstrapped model. 

This approach learns quickly from a device's encounters. On the other hand, when the data encountered is unbalanced with respect to the goal distribution, the model can overfit at the expense of the labels it encounters less frequently. Our second approach addresses this by computing a windowed average over a diverse set of recent encounters, where the diversity is determined by differences in the data label distributions of the contributed gradients. Every exchange with an encountered neighbor generates an update to $\Gamma$, which maps a neighbor's data label set to the  gradients learned from that set. During a new encounter, this approach averages over {\em all} of these stored gradients before generating the update to $w^{\prime}$. While this slows the speed of learning, it provides increased stability, especially when the learner encounters highly unbalanced data label distributions. We term this approach {\em opportunistic momentum}. Formally, the update to $w^{\prime}$ is computed as:
\begin{equation}
    \frac{\omega_{(\lab_i,\goal_i)}(\nabla\ell(w^{\prime};\mathcal{D}_i))+ \bigg(\sum_{(\lab,\nabla)\in\Gamma}\omega_{(\lab,\goal_i)} \nabla\bigg)}{\omega_{(\lab_i,\goal_i)} + \sum_{(\lab,\nabla)\in\Gamma} \omega_{(\lab,\goal_i)}}.
    \label{eq:oppMom}
\end{equation}
The first term of the numerator accounts for the (weighted) contribution of a round of training on $\device_i$'s local data. The second term sums the gradients stored in the non-empty entries in $\Gamma$, each weighted based on similarity. Before evaluating our approaches, we examine one final concept in our framework, the notion of decay with respect to the learning rate.

\subsection{Learning Rate and Decay}

Appropriately tuning the learning rate is important to avoid overfitting. We dynamically tune the learning rate by utilizing the concept of {\em decay}, which is common in deep learning~\cite{bengio2012practical}. Because our approach is completely decentralized, the learning rate for each device evolves independently. The learning rate at time $t_i$ for device $\device_i$ is $\eta^{t_i}_i = \eta\alpha^{t_i}_i$, where $\eta$ is an initial learning rate and $\alpha^{t_i}_i$ is the decay, computed as:
\begin{equation}
    \alpha_{t_i} = \frac{\exp(\kappa\times(\phi - \left\lVert w^0_i - w^{t_i}_i \right\rVert_2))}
    {\exp(\kappa\times(\phi - \left\lVert w^0_i - w^{t_i}_i \right\rVert_2)) + 1},
    \label{eq:decay}
\end{equation}
where
\begin{equation}
    \alpha_{t_i} < min\{\alpha_0, ... , \alpha_{t_i-1}\},\; 0 < \phi,\; 0 < \kappa,
\end{equation}
and $\phi$ and $\kappa$ are constants. Opportunistic federated learning avoids overfitting to continuous encounters with a heavily skewed dataset by making an assumption that the minimum for a personalized task exists somewhere not too far away from the bootstrap model on the loss surface. The decay factor $\alpha$ is a sigmoid function, where an L2 distance from the initial weight is scaled and used as an input. This design encourages $\device_i$'s model to find a solution near the bootstrap model, as we assume it ensures a certain level of performance for all labels. 

As the model becomes more personalized, the learning rate decreases proportionally to the decay factor to seek a more fine-grained solution. At the same time, we only take the minimum of the decay factor to prohibit the model from reverting completely back to the original solution (the bootstrap model). The values of $\eta$, $\phi$ and $\kappa$ are global constants determined prior to bootstrapping; like $\lambda$ above, we determined the values by running experimental training with a subset of training data used for bootstrapping the initial model.

%% file: empiricalSupport.tex
We benchmark and evaluate our opportunistic federated learning framework in two threads. First, we present controlled experiments in which we manipulate devices' encounter patterns and data distributions in order to learn about and demonstrate how these impact performance. We the use realistic scenarios to demonstrate how opportunistic federated learning might perform in more realistic scenarios.

\subsection{Datasets and Models}
Opportunistic federated learning requires training models on commodity mobile devices. Further, training must be completed within the timeframe of an encounter between two neighboring devices. For these reasons, the models most suitable for opportunistic federated learning are likely to be relatively small and lightweight tasks. Our evaluation relies on two classification tasks; MNIST and CIFAR-10. In future work, we will explore pushing opportunistic federated learning even more, with additional models and with models that grow in size and complexity. In MNIST~\cite{lecun1998gradient}, the task is to correctly label images of handwritten digits 0 through 9. We replicate a ``2NN'' model from~\cite{mcmahan2017communication}, which was used to prove centralized federated learning empirically. The network is composed of two fully-connected hidden layers, each with 200 neurons and ReLU activations. Our second dataset is CIFAR-10~\cite{krizhevsky2009learning}, where the task is to recognize objects in images. We use a convolutional neural network (CNN) model, which is 11 layers deep with convolutional, max pooling, and dropout layers. 
MNIST and CIFAR-10 have 60,000 and 50,000 training images respectively. We used 10\% and 25\% of the entire training set, respectively, to train the bootstrap model and used the remainder of the data to create the devices' local datasets.

We chose these datasets because (1)~they map to our motivating applications; (2)~the models can realistically be trained on resource-constrained devices and (3)~the datasets are sufficiently large. There are many applications that satisfy the first two constraints, but the third is more difficult to realize, in particular because our evaluation demands the ability to distribute the data in a skewed way among many devices.

\subsection{The Feasibility of Encounter-Driven Learning}
Opportunistic federated learning relies on coordinated rounds of device-to-device exchanges that occur when users' mobile devices encounter one another. It is essential to fit the execution of the exchange within the duration of an encounter. In particular, Lines 8-10 of Algorithm~\ref{alg:req} unfold as (1)~the learner sends the model; (2)~the neighbor performs one round of training; (3)~the neighbor returns the gradients; (4)~the learner performs one round of training and aggregation. These four steps are repeated $\rho$ times; the total needed time is:
\begin{equation}
    t_{\it enc} = \rho \times ( 2 t_{\it send} + 2 t_{\it train} + t_{\it agg}).\label{eq:time}
\end{equation}

To compute $t_{\it train}$ and $t_{\it agg}$, we measured the computation time on a Raspberry Pi 4 (which has computational capabilities comparable to a smartphone). Training  takes, on average, 1.543s and 5.74s for MNIST and CIFAR-10, respectively. For all approaches other than {\it opportunistic-momentum}, $t_{\it agg}$ is 0. In {\it opportunistic-momentum}, Line 10 of Algorithm~\ref{alg:req} requires iterating over the table $\Gamma$. The table has, at most, an entry for every subset of the goal set; however in practice the table is much smaller because it does not include entries that are completely subsumed by another and because a learner does not encounter all possible subsets of the goal set. To compute $t_{\it agg}$, we assumed the worst case $|\Gamma| = 2^{|\mathcal{G}|}$, or $|\Gamma|=32$, given that, in all of our experiments, the size of the goal distribution set is 5. Measured empirically on the Raspberry Pi 4, the worst case $t_{\it agg}$ for MNIST and CIFAR-10 are 0.064s and 0.448s, respectively, assuming $\Gamma$ is loaded in memory.

Finally, computing $t_{\it send}$ requires knowing the size of the model and the communication rate of the wireless channel; our MNIST  model is 778KB (199,210 parameters), while CIFAR-10 is 4.8MB (1,250,858 parameters). 
Assuming two devices are connected via WiFi-direct, whose datarate is 250Mbps, $t_{\it send}$ for MNIST is 0.020s and for CIFAR-10 is 0.153s. With a lower datarate Bluetooth connection (i.e., 2Mbps), the $t_{\it send}$ values are 3.05s and 19.1s, respectively.

\begin{table}[]
    \centering
    \caption{Required encounter durations ($|\Gamma|=32$, $\rho=6$).}
    \vspace{-.25cm}
    \begin{tabular}{|l|*{4}{p{0.8cm}|}}\hline
         & $t_{\it train}$ &  $t_{\it agg}$ & $t_{\it send}$  &  $t_{\it enc}$ \\\hline\hline
        MNIST$_{\rm WIFI}$ & 1.543s & 0.064s & 0.020s & 19.14s \\\hline 
        MNIST$_{\rm Bluetooth}$ & 1.543s & 0.064s & 3.05s & 55.50s \\\hline     
        CIFAR-10$_{\rm WIFI}$ & 5.740s & 0.448s & 0.153s & 73.40s \\\hline
        CIFAR-10$_{\rm Bluetooth}$ & 5.740s & 0.448s & 19.1s & 300.77s \\\hline
    \end{tabular}
    \label{tab:time}
\end{table}

Table~\ref{tab:time} shows $t_{\it enc}$ for both models. 
Many encounters in pervasive computing environments will satisfy these required durations, especially with a WiFi-direct connection (e.g., standing in line at the grocery store, chatting with a friend on the street, etc.). 
The longer durations needed when Bluetooth is used limit the usable encounters, but there are still many pervasive computing encounters that fall within this range (e.g., commuting on public transportation, eating in a restaurant, or sitting in a meeting). For our simulations we use a datarate of 1Mbps (a not-quite-ideal Bluetooth connection).

\subsection{Evaluation Platform}
We implemented our framework and algorithms in Python using TensorFlow~\cite{tensorflow2015-whitepaper} and Keras~\cite{chollet2015keras}.\footnote{\url{https://github.com/UT-MPC/swarm}} The models can run on resource constrained mobile devices. For the purposes of this paper, we also created a simulation environment that simulates each device's instance of the framework separately. The simulation environment provides an implementation of the ``device-to-device'' communication by passing messages between the threads. It also simulates the contact patterns that drive the encounters between the simulated devices. 

Determining whether to engage in an exchange has two components: (1)~whether it is likely beneficial, based on the similarity between the data label and goal distributions and (2)~whether it is feasible based on the predicted encounter duration. For the former, we use the similarity as computed in Equation~\ref{eq:sim}. For the latter, we assume that predictions of contact duration from the underlying system are perfect; relaxing this assumption is left for future work. Given the predicted duration, we use Equation~\ref{eq:time} for the amount of time required to complete an encounter.

\subsection{Controlled Experiments}
In our first experiments, we tightly controlled encounters and data distributions so that we could carefully benchmark our framework. We performed extensive evaluations on both datasets; due to space constraints, we show just one example. We compare the performance of five approaches:
\begin{itemize}
    \item {\em local}: the model trains only on the learner's local data; for comparison purposes, the model continues to train over time even though no new data is generated.
    \item {\em pairwise-fed-avg}: the model trains using a pairwise version of federated averaging~\cite{konevcny2016federated}; a pair of devices perform as many rounds of pairwise federated averaging as each encounter duration allows, starting with a base model that is the average of the two devices' models, rather than from the learner's model as in the remaining approaches.
    \item {\em greedy-no-sim}: the model trains on every encounter without considering the similarity between the learner's goal distribution and the neighbor's data label distribution using the greedy aggregation from Equation~\ref{eq:greedy} with $\omega = 0.5$ (i.e., equal weight to  local data and neighbor's data).
    \item {\em greedy-sim}: we limit the training encounters to those of sufficient similarity ($\tau=0.2)$, still using $\omega = 0.5$.
    \item {\em opportunistic-momentum}: we train the model using the strategy in Equation~\ref{eq:oppMom} with $\tau=0.2$, including weights determined according to Equation~\ref{eq:weights} and dynamic decay of the learning rate (Equation~\ref{eq:decay}).
\end{itemize}
The first two models are baselines for comparison; the remaining three models are all novel contributions of our work. Our goal is to understand the conditions under which each is suitable for supporting opportunistic federated learning.

\begin{figure}[t]
\centering
\begin{subfigure}[l]{\columnwidth}
\includegraphics[width=\columnwidth]{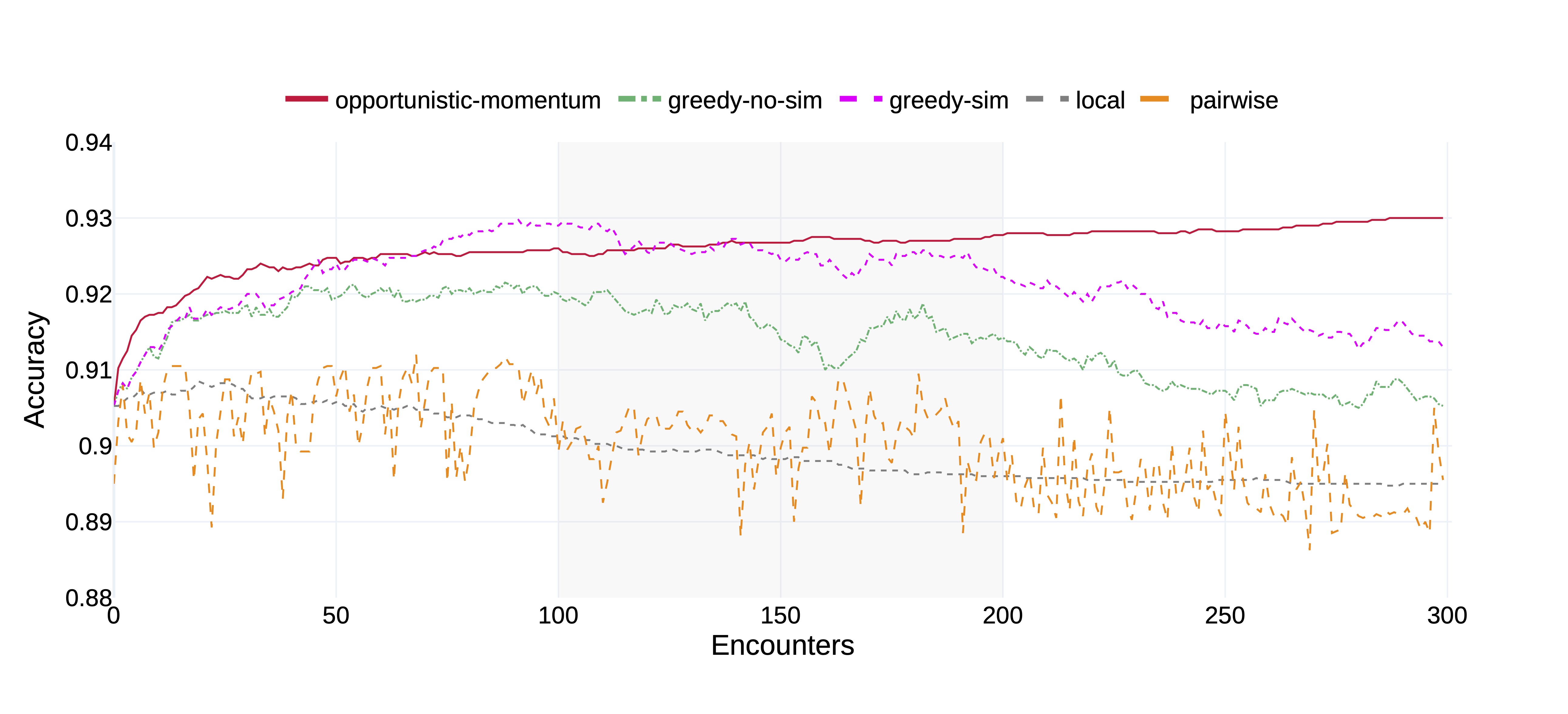}
\caption{\small MNIST}
\label{fig:mnistcontrolled}
\end{subfigure}
\begin{subfigure}[l]{\columnwidth}
\vspace{.2cm}
\includegraphics[width=\columnwidth]{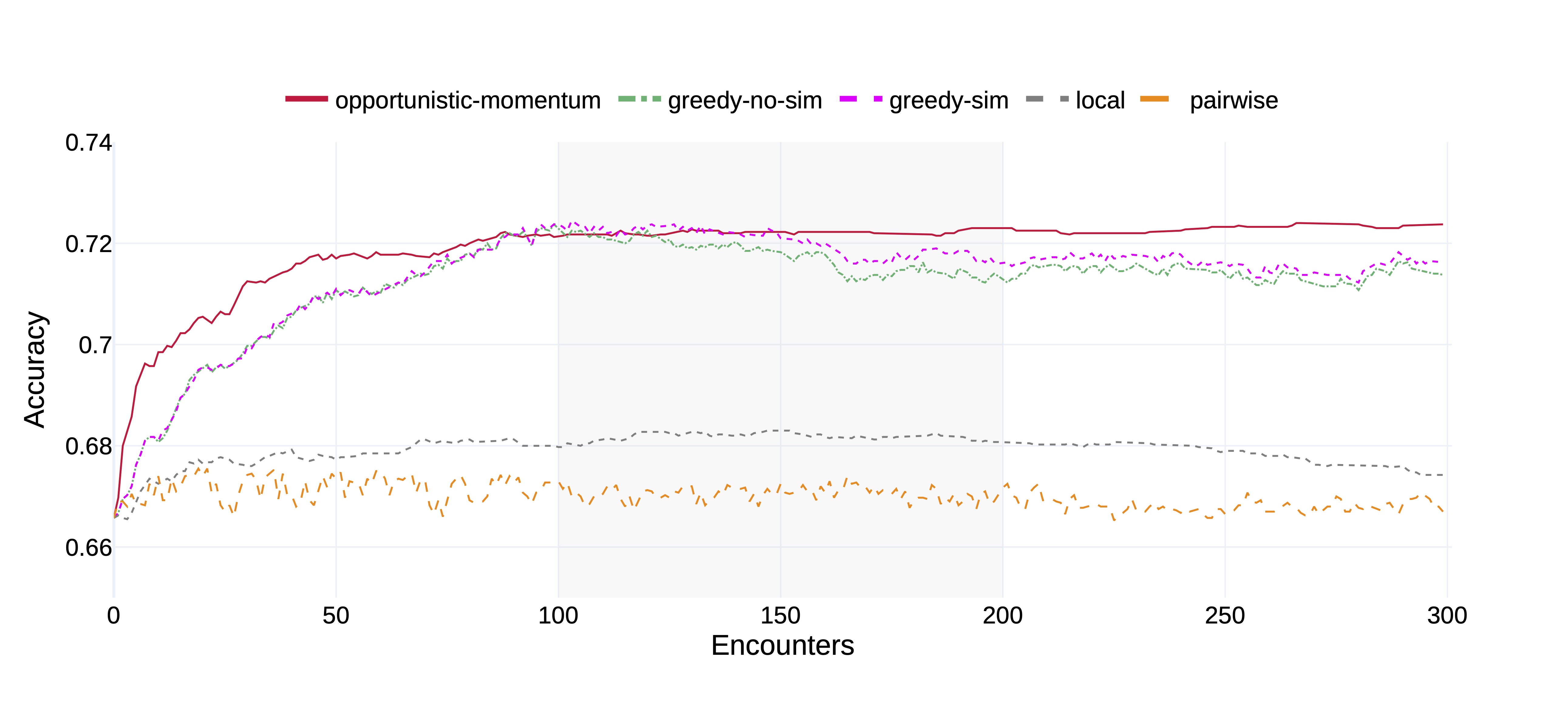}
\caption{\small CIFAR-10}
\label{fig:cifarcontrolled}
\end{subfigure}
\caption{\small Comparing algorithm options for (a) MNIST and (b) CIFAR-10. The x-axis is the number of encounters the device has experienced. The y-axis is the model accuracy. 
Each shaded region is a shift in the data distributions on the encountered devices.} 
\label{fig:controlled}
\end{figure}

Fig.~\ref{fig:controlled} shows results for both datasets. MNIST has a label for each digit in the range 0-9; in CIFAR-10 there is a label for each class of object recognized in a photo. For simplicity, we refer to both label sets with numbers 0-9. In Fig.~\ref{fig:controlled}, the goal distribution contains exactly five of the ten labels, specifically labels $\{0, 1, 2, 3, 4\}$. Every device has a local dataset of with 80 items in MNIST and 150 in CIFAR-10. The learner's local dataset contains equal numbers of labels $\{0,1\}$. The encountered data label distributions change over time:
\begin{itemize}
    \item {\em first 100 encounters}: 50\% of the first 100 encounters are with neighbors that see exactly and only labels $\{2, 3\}$, and the other 50\% are with neighbors that see three labels, selected randomly from all ten labels.
    \item {\em encounters 100-200}: in the middle, 50\% of encounters are with neighbors that see exactly and only labels $\{3, 4, 5\}$, and the other 50\%  are with neighbors that see only three labels, selected randomly from all ten labels.
    \item {\em encounters 200-300}: in the last period, 50\% of encounters are with neighbors that see exactly labels $\{4, 5, 6\}$, and the other 50\% are with neighbors that see only three labels, selected randomly from all ten labels.
\end{itemize}

A completely egocentric approach that trains only on a device's local data ({\em local} in Fig.~\ref{fig:mnistcontrolled}) results in a model that overfits very quickly. Pairwise federated averaging ({\em pairwise-fed-avg}), while intuitively promising, also suffers under these workloads (and in real environments). The reason is that federated averaging combines the models of the two devices to generate a new base model used for training. Because these devices have been working independently, their models have likely diverged. In contrast, the remaining approaches all use the learner's model as the base, even on the neighbor's device.

The next approaches are {\em greedy-no-sim} and {\em greedy-sim}. The former greedily incorporates whatever it can achieve with any encountered neighbor, without considering its goal distribution. In this particular example, {\em greedy-no-sim} performs the worst of our approaches because this example has a very unbalanced and unstable distribution of encountered data. When the encountered data is more well mixed, {\em greedy-no-sim} performs better and is, in some cases, the best performing approach. The downside of {\em greedy-no-sim} is most apparent in the third region, when the likelihood that the device encounters data label distributions that overlap with its goal distribution decreases. In contrast, {\em greedy-sim} is more resilient to changes in encountered data label distributions because it only requests learning from devices whose distributions are sufficiently similar to the goal distribution. This example uses a value of $\tau=0.2$ in Algorithm~\ref{alg:req}; even this minimal overlap has a substantial benefit to performance. Larger values for $\tau$ result in even better performance for {\em greedy-sim}, albeit in exchange for somewhat slower convergence.

In this example, {\em opportunistic-momentum} ended with the highest accuracy. {\em Opportunistic-momentum} is the most resilient to dramatic changes in data distributions, and it is best at avoiding overfitting to a distribution it sees for a period of time because it continuously integrates meaningful gradients that it previously collected, even if it does not continue to encounter them. 

The greedy approaches dip in performance as we move from one region to another because the models have overfit to the data label distributions. In contrast, {\em opportunistic-momentum} is quite resilient to sudden changes in data label distribution. Arguably, this is a contrived situation, designed to benefit {\em opportunistic-momentum} relative to other strategies. If the data distributions that the user encounters are highly overlapping with the goal distribution, the greedy strategies outperform {\em opportunistic-momentum}. For this reason, we now step into more real-world experiments, where the data distributions are much less controlled and contrived. 

\subsection{Real World Scenarios}
We next allow devices' movement patterns to evolve independently according to the Levy walk mobility model~\cite{birand2011dynamic, rhee11:on}, which, among stochastic mobility models, is known to capture human mobility well~\cite{birand2011dynamic, hong2008routing}. Levy walk assumes that the majority of an individual's movements are in a small local area, with a few very large movements every once in a while.
 
We opted for randomized mobility that allows more careful understanding of the impacts of mobility, which in turn enables a more careful benchmarking of the performance of our algorithm. Future work will include evaluation on collected mobility traces and on real devices.

We created a square space, divided into 9 equally sized regions. Each device is assigned an anchor region (e.g., the user's home), with five devices assigned to each region. 

We simulate {\em episodes}; in each episode, the user starts out at home, makes some trips away from home, and returns home at the end of the episode. Each runs consists of 10 episodes. Devices encounter one another by coming within a pre-defined communication range; some encounters are ephemeral, as the devices move past each other on their trajectories, while others are longer-lasting, as devices stay nearby for some period of time. We assume that devices have perfect knowledge of predicted encounter durations.

Each region is associated with two labels. Any device's whose anchor location is in that region has a data label distribution that contains exactly those two labels. Each device sets its goal distribution to be those two labels and three additional randomly chosen labels. As such, devices with anchor locations in the same region likely have different goal distributions. This distribution of data and goals  mimics a real world environment where devices collect different data depending on their experiences but also have diverse goals, e.g., due to the fact that their different travel patterns lead them to encounter different data that needs to be classified.

\begin{figure}[t]
    \vspace{-6mm}

    \begin{subfigure}[b]{0.49\columnwidth}
        \includegraphics[width=\columnwidth]{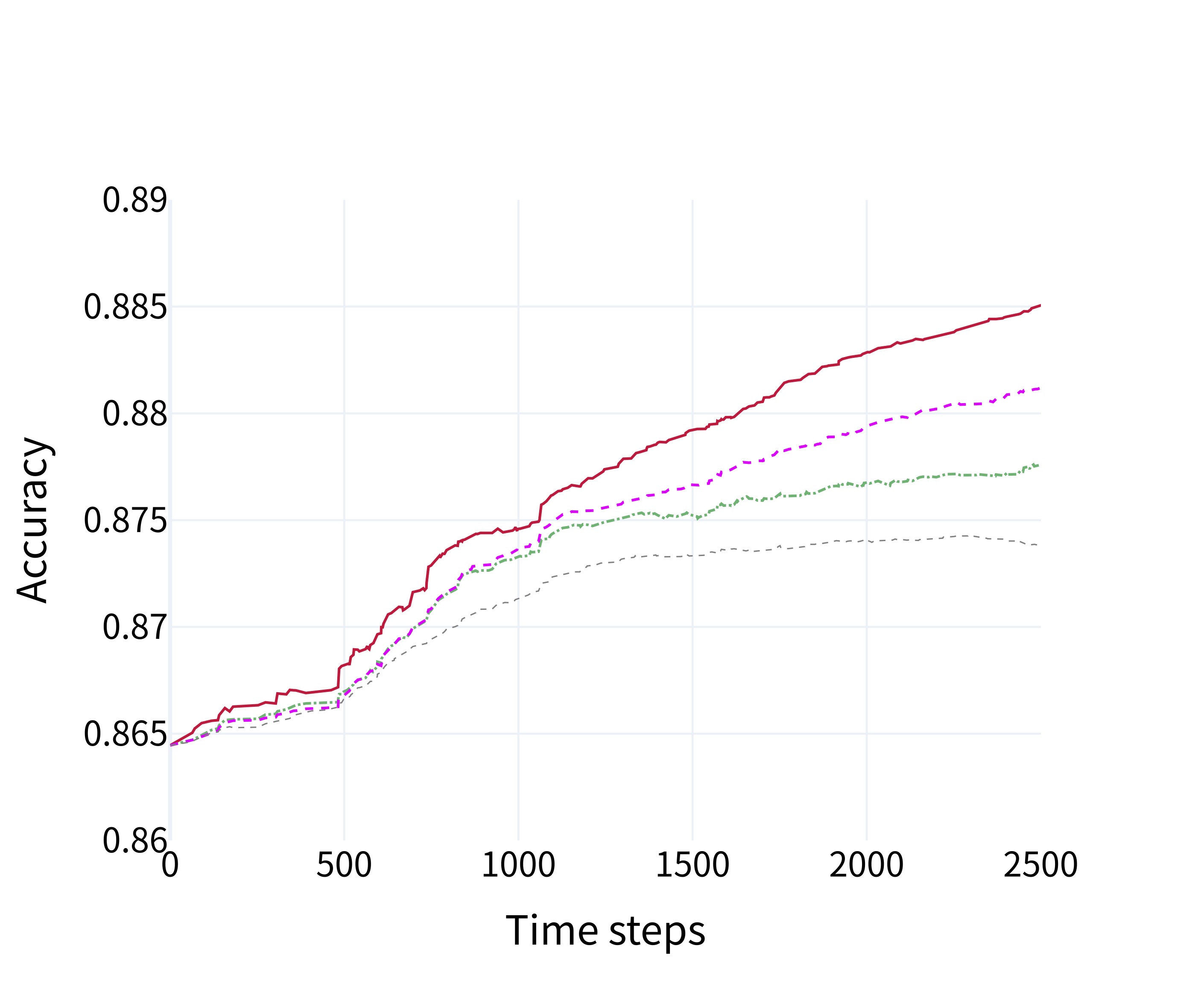}
    \end{subfigure}
    \hfill
    \begin{subfigure}[b]{0.49\columnwidth}
        \includegraphics[width=\columnwidth]{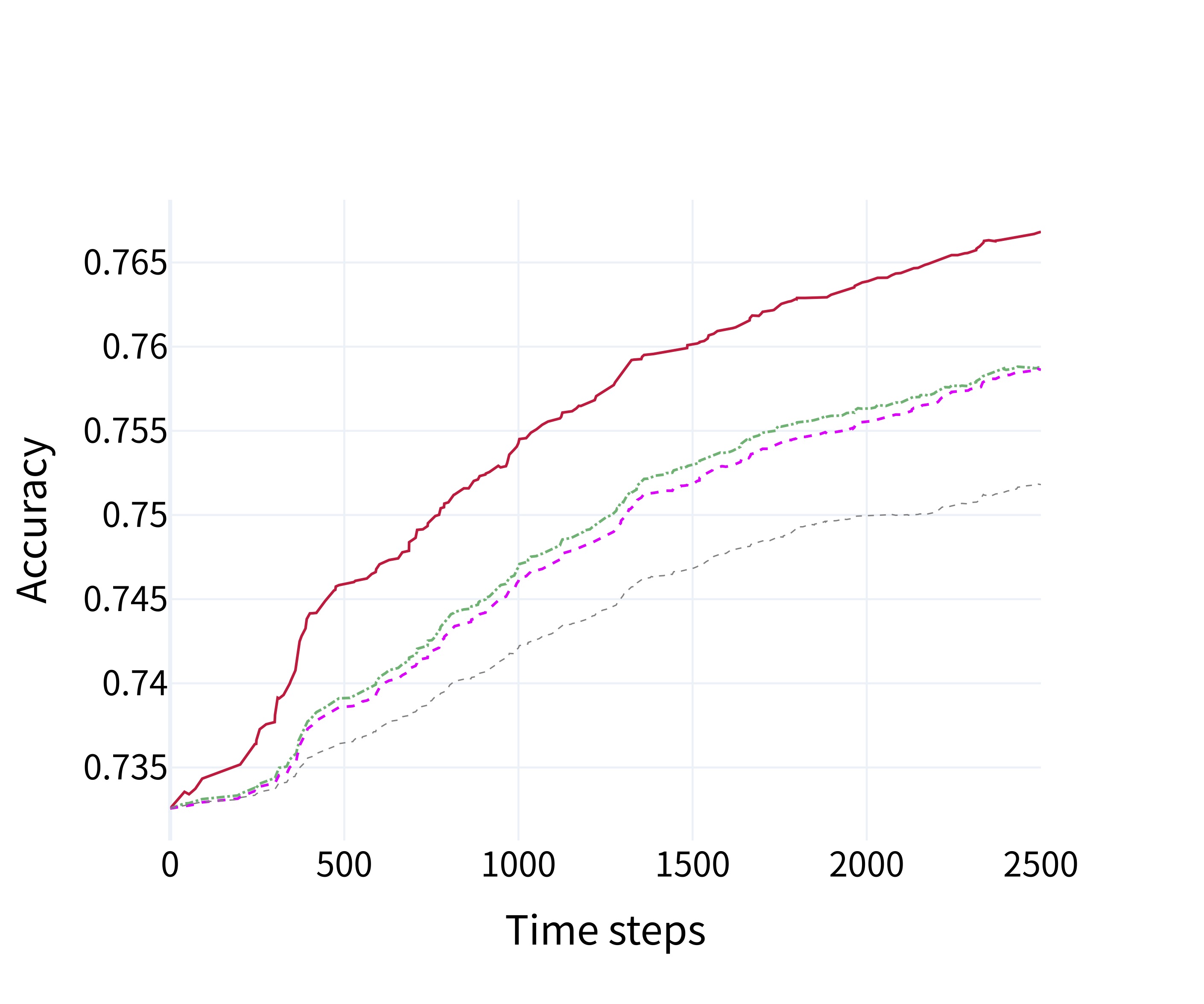}
    \end{subfigure}

    \begin{subfigure}[t]{0.9\columnwidth}
        \hspace{5mm}\includegraphics[width=\columnwidth]{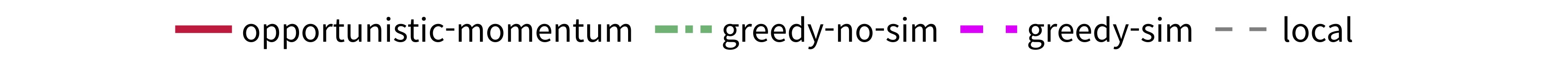}
    \end{subfigure}
  \caption{\small Comparing algorithm performance for a real-world scenario for (a) MNIST and (b) CIFAR-10. The x-axis is the elapsed time of the simulation. The y-axis is the model accuracy.}
   \label{fig:final-sim}
\end{figure}

Fig.~\ref{fig:final-sim} shows the accuracy averaged over all 45 devices. For both models, the framework performs as expected, given the results in the controlled experiments. For CIFAR-10, however, {\em greedy-no-sim} slightly outperforms {\em greedy-sim}.

In this scenario, the data label distributions and goal distributions have a significant random component to them. As a result, the labels are relatively well distributed among devices, putting our models in a situation similar to the far left region of Fig.~\ref{fig:controlled}.  This benefit does come at a cost; because {\em greedy-no-sim} takes advantage of every possible encounter, it has 18\% higher overhead compared to {\em greedy-sim}.

This last observation opens a piece of future work: designing an approach that can adapt to the changing nature of surrounding data distributions. When the encountered data is evenly distributed, the algorithm can enter a {\em greedy-no-sim} mode, taking advantage of any and all opportunity for collaboration. When the algorithm senses an imbalance in encountered data, it could transition into {\em greedy-sim}.

In conclusion, these results show that our approaches to opportunistic federated learning can consistently outperform local personalized learning and can be very resilient to overfitting and dramatic fluctuations in the encountered data distributions.

%% file: conclusion.tex
This paper explored a novel paradigm for machine learning in pervasive computing, which we term {\em opportunistic federated learning}. We defined a framework through which devices can collaborate opportunistically with other devices in their surroundings using only device-to-device communication links. We defined an algorithm within the framework, {\em opportunistic momentum}, which provides a robust mechanism to continuously integrate learning from encounters in a way that improves over a well-informed greedy approach. Overall, our results demonstrate that there are real-world pervasive computing scenarios and applications that can garner significant benefits from this collaborative yet personalized approach to {\em in situ} training of reasonably coupled deep learning models. 

%% file: futureworks.tex

Our results show the significant promise of opportunistic federated learning for diverse pervasive computing applications. Near term future work will extend our evaluation to even further understand the performance of the opportunistic-momentum algorithm, including assessing the impacts of increased varieties of goal distributions, evaluating even more models (e.g., human activity recognition models), and relying on real-world mobility traces. We should also consider what happens when a device's goal distribution changes, either gradually (e.g., as emoji trends come and go) or abruptly (e.g., because a user changes their job or home environment). It is possible that our approach will successfully adapt to gradual changes, but more abrupt changes will require re-bootstrapping a model, perhaps from an encountered neighbor.